\documentclass[letterpaper]{article} 

\usepackage{aaai2027}  

\usepackage[hyphens]{url}  

\usepackage{graphicx} 
\usepackage{natbib}  
\usepackage{caption} 
\usepackage{algorithm}
\usepackage{algorithmic}
\usepackage{array}

\usepackage{newfloat}
\usepackage{multirow}
\usepackage[utf8]{inputenc}
\usepackage{textgreek}
\usepackage{listings}
\DeclareCaptionStyle{ruled}{labelfont=normalfont,labelsep=colon,strut=off} 
\floatstyle{ruled}
\newfloat{listing}{tb}{lst}{}
\floatname{listing}{Listing}

\usepackage{booktabs}

\title{P-ContextBench: From Personal Fact Recall to Personalized Agent Behavior}
\author{
    Grant Nasir Brown,
    Abdul Basir Bari,
    Anmol Tadikonda,
    Isabel Song
}

\affiliations{
    Department of Computer Science,
    San Francisco State University\\
    gbrown8@sfsu.edu, abari@sfsu.edu,
    atadikonda@sfsu.edu, hjsong@sfsu.edu
}

\begin{document}

\maketitle

\begin{abstract}
Large language models (LLMs) are increasingly used for personal tasks, yet evaluating whether they can safely use personal context remains difficult because such contexts are private, fragmented, dynamic, and socially situated. We introduce P-ContextBench, a realistic, verifiable, and scalable benchmark framework for evaluating personal-context understanding. P-ContextBench constructs population-structured, life-story-grounded digital humans from demographic seeds reflecting the U.S. population. Rather than presenting personal context as explicit and static profiles, P-ContextBench embeds traits, routines, goals, and constraints as implicit evidence distributed across fragmented and evolving synthetic mobile-app records.  It evaluates five capabilities: single-source fact recovery, multi-source fact recovery, temporal state tracking, inference of higher-level personal patterns from routine behavior, and safe handling of user requests that conflict with persona-specific constraints—for example, by complying, warning, clarifying, or refusing. A multi-stage validation pipeline checks persona plausibility, recoverability of the intended personal context from the evidence, information leakage, evidence preservation, test-case grounding, and human–judge agreement. We instantiate P-ContextBench with 100 personas, 10,088 hidden facts, 18,261 app-trace fragments, and 4,810 verified test cases, while the benchmark pipeline supports configurable generation at different scales and for different demographic groups. Cost-effective state-of-the-art (SOTA) LLMs relevant to personal-agent settings achieve below 60\% accuracy on temporal tracking, higher-level pattern inference, and constraint-aware action-boundary decisions, with less than 10\% of test cases answered correctly in all three repeated runs. Even a high-performing SOTA model remains inconsistent on these capabilities. P-ContextBench thus provides measurable targets for improving the accuracy, reliability, and safety of LLM-based personal agents. Code and data are available at https://github.com/TruePersona/P-ContextBench.
 
\end{abstract}

\section{Introduction}

In recent years, large language models (LLMs) have increasingly been deployed to assist users with personal, everyday tasks, such as shopping, scheduling appointments, and making daily decisions.  In these consumer-oriented settings, personalization is a critical requirement. LLMs must adapt to a user’s specific context, including preferences, past events, personal traits, goals, and constraints. Effective personalization requires more than fact recovery or following an immediate user request; it requires applying personal context to produce useful, appropriate, and safe responses. This capability becomes increasingly important as LLMs are embedded deeply in everyday life. 

	Personal-context benchmarking introduces challenges not captured by traditional LLM evaluations. Personal context is rarely available as a consolidated, static profile; instead, it is fragmented across multiple sensitive digital traces such as app logs, which are difficult to access, collect, or share ethically because of privacy concerns. Personal context is often revealed implicitly through behavior rather than direct statements, and it changes over time as preferences, constraints, routines, relationships, and goals evolve. As LLMs are increasingly used for a growing range of everyday tasks, personalization must move beyond the simple fact recall emphasized in existing benchmarks. It requires interpreting how multiple facts jointly shape a person’s needs, constraints, and appropriate actions. Hence, models need to synthesize low-level details into higher-level personal patterns, such as motives, tradeoffs, constraints, and decision rules, and determine how to act when user requests involve conflicting goals, ambiguity, or exceptions. Since personal task patterns vary across individuals and demographic groups, and app usage trends evolve rapidly, personal-context benchmarks should also be extensible across populations and different app sources.

  To address these challenges, we propose \textbf{P-ContextBench}, a realistic, verifiable, and scalable benchmark framework. Unlike existing benchmarks that assign predefined traits to a persona, P-ContextBench first constructs a population-structured life-story-grounded digital human. It starts from demographic seeds  reflecting the U.S. population and enriches each seed through a comprehensive interview of over 100 questions covering personal history, family and social relationships, work, health, finances, values, daily routines, and views on social issues, thereby forming a coherent representation of an individual. P-ContextBench then extracts hidden facts—personal traits, routines, goals, and constraints used as benchmark ground truth—from this digital human, rather than assigning them as isolated attributes, thereby grounding downstream app traces in plausible personal context.  Building on these facts, P-ContextBench synthesizes each persona’s social circle and longitudinal app traces, which implicitly represent personal context as fragmented, socially situated, and temporally distributed over time without using real user data. The benchmark construction pipeline is scalable: demographic seed selection can target specific populations, and app-agnostic hidden facts and social circles can be instantiated as different app-log formats. Each applicable construction stage is checked by a separate LLM for consistency, recoverability, leakage control, and grounding.  

  	To evaluate LLMs’ personal intelligence in an everyday setting, P-ContextBench includes five task types: \textbf{single-source fact recovery (T1), multi-source fact recovery (T2), temporal state tracking (T3), higher-level personal-pattern inference  (T4)}, and \textbf{constraint-aware agent behavior (T5)}. T1–T4  test whether models can recover, connect, update, and reason over fragmented personal context. T5 tests whether models can set appropriate action boundaries when user requests conflict with persona-specific constraints, such as complying, warning, clarifying, suggesting alternatives, or refusing. For example, if a user with diabetes requests to buy chocolate, the model must recognize the health constraint and decide whether to warn, clarify, or suggest an alternative rather than simply comply.    
    
	We provide an implementation that supports configurable benchmark generation at different scales and for different demographic groups. For evaluation, we instantiate P-ContextBench with 100 personas, 500 social-circle contacts, 10,088 hidden facts, and 4,810 test cases. Personas are verified for demographic plausibility, behavioral consistency,  psychological coherence and political consistency. App logs are checked to ensure that hidden facts are implicitly encoded and recoverable from the supporting evidence. Test cases are verified to ensure that their answer keys are grounded in the hidden facts, supported by the app logs, and valid for their intended task types. The scoring protocol is cross-checked by human annotators. 
  
	On P-ContextBench, we evaluate cost-effective state-of-the-art (SOTA) models relevant to personal-task settings, along with a high-performing reference model. We report overall, worst-run and Pass$^{3}$ accuracy \cite{shi2026tau}, where Pass$^{3}$ counts a test case as correct 
    only when the model receives full credit in all three independent runs. The cost-effective models achieve 48–62\% overall accuracy, but  less than 20\% Pass$^{3}$, with Pass$^{3}$ below 6\% on each of T3, T4 and T5. This gap shows that models often succeed inconsistently across repeated runs.  Even the high-performing reference model exhibits a persistent reliability gap, achieving 75.52\% overall accuracy but only 34.65\% Pass$^{3}$
    on average, underscoring the challenge posed by  P-ContextBench. 
    
	Failure analysis shows a common pattern: models often find relevant evidence but fail to use it   completely. In  T3, 81.01\% of non-correct cases omit a required current-state hidden fact. In T4, 64.78\% omit hidden facts needed to support the reasoning conclusion. In T5, failures center on decision quality, led by constraint-recognition errors (37.53\%) and over-refusal (13.62\%). Together, these findings reveal concrete targets for improving the reasoning, reliability, and safety of LLM-based personal agents. 
    
	To our knowledge, P-ContextBench is the first personal-context benchmark framework to combine population-structured digital-human personas, tasks spanning fact recovery through personalized agent behavior, and scalability across populations and app formats. Our contributions are: 

\begin{itemize}  
\item \textbf{A realistic and scalable benchmark framework.} P-ContextBench represents personal context through coherent digital-human personas, social circles, hidden facts and app traces; evaluates diverse tasks; and supports configurable generation across different scales and demographic groups.
\item \textbf{A verifiable benchmark construction pipeline.} We instantiate P-ContextBench with 100 personas, 500 social-circle contacts, 10,088 hidden facts, 11,930 messenger fragments, 6,331 calendar fragments and 4,810 test cases, with stage-wise verification.
\item \textbf{An empirical analysis of LLM limitations.} We identify key gaps in temporal updating, higher-level personal-pattern inference, constraint-aware action-boundary decisions and reliability, providing concrete improvement targets for LLMs and personal-agent systems.
\end{itemize}

Code and data are available at https://github.com/TruePersona/P-ContextBench.

\section{Related Work}
Prior work has explored persona synthesis and associated digital traces, while a growing body of research has developed personal-context benchmarks. 

\subsection {Synthetic Data Generation for User Modeling}
Park \cite {park} introduced a life-story-grounded digital human simulation framework for modeling real individuals. They recruited 1,052 U.S. participants, conducted qualitative interviews covering over 100 questions, and conditioned LLM agents on those interview transcripts to simulate each participant across social, attitudinal, personality, and decision-making contexts. The resulting digital humans demonstrated high fidelity to real individuals, achieving 0.85 normalized accuracy on the General Social Survey, measuring demographics and socio-political attitudes,  and 0.80 correlation on the 44-item Big Five Personality Inventory, measuring core personality traits. The results show that the simulation framework is a strong basis for human modeling. Ge \cite {ge2024scaling} proposed PersonaHub, a persona-driven synthetic data generation framework that uses personas to guide the creation of diverse synthetic assets, including knowledge-rich texts, game NPCs, and social connections, with a reported expert validity rate of 96.5\%.

\subsection {Personal-Context Benchmarks}  
Several benchmarks evaluate LLMs in personal-context settings, especially whether models can infer user preferences over time. PersonaMem \cite{jiang2025know} simulates user--chatbot histories and shows that SOTA LLMs struggle with dynamic preference shifts even when preferences are explicitly represented. PersonaMem-v2 \cite{jiang2025personamem} extended this to implicit personalization, where preferences, traits, or constraints are revealed indirectly across up to 10  user--chatbot sessions. It shows even frontier LLMs remain limited in inferring user preferences from subtle signals in long histories. However, its primary evaluation is targeted for preference-centered personalization rather than broader daily-life personal-context use. 

 PersonaBench \cite{tan2025personabench} broadens the evaluation setting by synthesizing private user data beyond user--chatbot histories, including messenger chats, social circles, and  purchase histories. It evaluates mainly on information retrieval over a RAG pipeline \cite{rag}; reasoning tests are limited to relationship-grounded personal context and comprise about 10\% of the benchmark. LoCoMo \cite{maharana2024evaluating} and LongMemEval \cite{wu2024longmemeval} are related to personal-context benchmarking because they evaluate long-term memory over multi-session interactions. LoCoMo uses up to 35 user–user conversation sessions, while LongMemEval covers up to 500 user–assistant sessions. Both test recall, aggregation, temporal reasoning, knowledge updates, and abstention, showing that SOTA LLMs still struggle with long-term conversational memory. 
 
	Recent work also evaluates personalized LLM agents in conversational and tool-augmented settings. PersonaLens \cite{zhao2025personalens} measures task completion, personalization quality, and response naturalness in simulated task-oriented dialogues without actual tool invocation. ETAPP \cite{hao2025evaluating} evaluates personalized tool use in a sandbox with 33 APIs, testing whether LLM agents select tools and fill parameters in accordance with user profiles. However, these benchmarks assume that the requested task is valid and should be completed. For  personal agents, a more fundamental challenge is deciding whether an action should be taken at all under persona-specific constraints before tool use or task execution begins. 
    
	Related work in enterprise and agent safety evaluates whether LLM agents comply with explicit organizational policies, guidelines, or safety constraints during task execution. Benchmarks such as τ-bench \cite{yao2024tau}, $\tau$-knowledge \cite{shi2026tau}, GuideBench \cite{diao2025guidebench}, Agent-SafetyBench \cite{zhang2024agent}, and R-Judge \cite{yuan2024r} test policy following, guideline compliance, and risk awareness. These works focus on externally specified operational constraints, whereas personal agents should often infer constraints from personal context that are implicit, fragmented, and evolving. 
    
	Existing personalization benchmarks have made important progress in evaluating personalized intelligence, but several limitations remain. First, they rely on synthesized digital assets, such as user--chatbot conversations or app logs involving assigned preference annotations, yet provide limited justification for how these assets represent coherent individuals,  realistic populations, and realistic daily-life contexts. Second, most benchmarks primarily emphasize fact retrieval. Although some include reasoning tasks, they do not comprehensively cover broader forms of personal-context use, including 
    motive, identity, tradeoff, coping-strategy, and decision-rule inference.
    They also do not test action-boundary decisions, where an agent must determine whether and how to act before taking action, especially when a user request conflicts with persona-specific constraints. Third, many benchmarks lack stage-wise verification of the full construction process, including the underlying persona data, which serves as the foundation for downstream traces and benchmark test cases. Finally, existing datasets are not designed to be scalable or configurable across different demographic subsets or future app environments, even though different populations and app environments may produce different personal-context patterns.

\section{P-ContextBench}
P-ContextBench is built around three design goals: realism, verifiability, and scalability. First, it models persona  and personal context as they  appear in everyday life: population-structured personas with broad lifestories,  and fragmented, implicit, dynamic, and socially situated context across multiple app traces. Its test cases cover a broad range of personal-context capabilities, including factual recall,  temporal state tracking, higher-level personal-pattern inference, and action-boundary decisions.  Second, it makes the benchmark verifiable across the applicable stages of the pipeline by checking persona plausibility, hidden-fact recoverability, leakage control, evidence preservation, test-case grounding, and scoring validity. Third, it is designed to scale across different populations and  app formats. 

\begin{figure*}[!t]
\centering
\includegraphics[width=\textwidth]{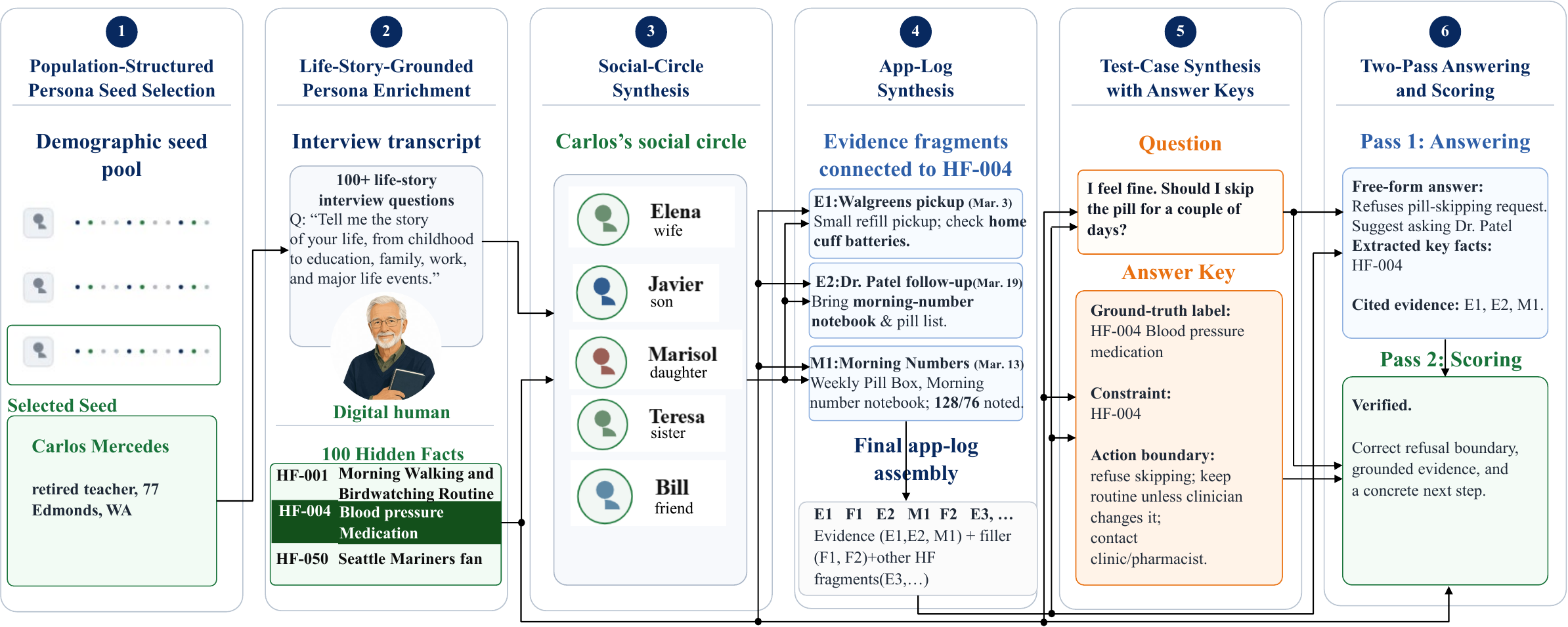}

\caption{Overall pipeline of P-ContextBench. $E_x$, $M_x$, and $F_x$ denote calendar, messenger, and filler fragments.}
\label{pipeline}
\end{figure*}

P-ContextBench is constructed through a six-stage pipeline as shown in Figure~\ref{pipeline}: (1) population-structured persona seed selection, (2) life-story-grounded persona enrichment through interviews, (3) social-circle synthesis, (4) app-log synthesis, (5) test-case synthesis with answer keys, and (6) two-pass answering and scoring. In the final stage, the model first answers given the app logs, and a separate scorer LLM grades each answer using the answer key, against type-specific rubrics.  
\subsection{Persona, Social-Circle and App-Log Synthesis}
The P-ContextBench  pipeline begins by creating a coherent digital-human from demographic seeds—brief profiles specifying attributes such as age, occupation, location, household background, hobbies, and career goals—sampled from NVIDIA Nemotron-Personas-USA \cite{nvidia}, which is grounded in demographic distributions from the U.S. Census \cite{uscensusdata}. P-ContextBench can sample seeds randomly for broad coverage or filter them by attributes such as age or occupation to target specific subpopulations.  

Each seed is expanded through a structured life-story interview of over 100 questions, following prior work \cite{park}, to construct coherent digital humans.  The interview responses are distilled into a hidden-fact inventory across 18 categories, where each hidden fact contains a canonical ground-truth label that is used in downstream benchmark construction and evaluation. A four-perspective LLM audit checks hidden facts for demographic plausibility, behavioral consistency, psychological coherence and political consistency.
    
Next, inspired by PersonaHub \cite{ge2024scaling}, we synthesize the persona’s close social circle, consisting of a fixed number of frequent contacts. Each contact is grounded in the interview transcript and hidden facts, including relationship history and shared activities. A verification step removes unsupported or contradictory details.
\subsection{App-Log Synthesis}
The fourth stage converts each persona's hidden-fact inventory and social circle into longitudinal app logs over a configurable time window. We instantiate two app-log sources: a messenger app log containing conversations with social-circle contacts and a calendar app log containing scheduled events.
\subsubsection{Implicit, Fragmented, and Leakage-Controlled Encoding}
To mirror real app usage, related hidden facts, typically two or three, are grouped into small clusters when they could naturally co-occur within the same routine, relationship, appointment, payment cycle, or planning context. For each cluster, the pipeline generates 4--5 app-specific fragments, distributed across app formats such as messenger sessions and calendar events.
These fragments encode hidden facts through behavioral, logistical, routine, and social signals, such as pharmacy pickups, copays, or recurring calendar entries, rather than directly stating their ground-truth labels, such as a disease name. Each hidden fact is supported by at least two fragments, so recovery requires combining multiple signals rather than reading a single explicit line. This is exemplified in Figure~ \ref{pipeline}. 
\subsubsection{Recoverability and Leakage Check}
After fragment generation, a separate LLM verifies that hidden facts are inferable without being directly leaked. For each hidden-fact cluster, the verifier sees only the corresponding 4--5  fragments, with metadata such as ground-truth labels and hidden fact IDs removed, and attempts to recover the intended hidden facts. It reports confidence for each inference and flags any direct label leakage, close synonym, or overly explicit cue. The generated fragment set is retained only when the intended hidden facts are recovered with high confidence and no leakage is detected; otherwise, it is regenerated.  
\subsubsection{Final App-Log Assembly}
To make the final app logs realistic, we add filler around the verified app-log fragments. Filler consists of ordinary non-evidence messages and calendar events, such as errands, reminders,  meals, and casual scheduling. 
Its amount is controlled by a configurable filler to-evidence ratio, measured as the number of filler tokens per meaningful evidence token.
To add realistic distraction, a configurable portion of the filler, from 0\% to 30\%, is reserved for decoy sessions. These decoys appear relevant to hidden facts but refer to someone other than the persona. A separate LLM verifies both that all verified evidence fragments are preserved in the final app log and that any overlap between a decoy and a hidden fact clearly belongs to another person, preventing the decoy from becoming accidental evidence. 
This process produces realistic but verifiable final app logs, where hidden facts are distributed across multiple layers of everyday app activity. Although the current benchmark uses messenger and calendar logs, its app-agnostic hidden facts and social-circle structures make the pipeline scalable to additional app formats.


\subsection{Benchmark Test-Case Synthesis}
Beyond simple information retrieval, P-ContextBench evaluates whether LLMs can recover nuanced personal context, infer higher-level personal patterns, and apply that context in realistic personal-agent situations such as respecting personal constraints and deciding when to act or abstain. We synthesize test cases grounded in each persona’s hidden-fact inventory and verified app logs,   emphasizing inference, judgment, and action-boundary reasoning.  P-ContextBench includes five task types: T1 single-source fact recovery, T2 multi-source fact recovery, T3 temporal state  tracking, T4 high-level personal-pattern inference, and T5 agent-oriented behavior. T1 recovers a fact from one app source while T2 connects evidence across messenger and calendar logs to recover a fact.  T3 requires models to distinguish prior state, current state, and stale interpretations; T4 requires inference of implicit personal patterns such as motives, identities, and values; and T5 requires models to infer persona-specific constraints and decide whether to comply, clarify, suggest an alternative, require confirmation, or refuse. T1–T4 use third-person questions about the persona, whereas T5 uses first-person requests to the agent.
For each test case, the pipeline creates a question and an answer key containing the required hidden facts, represented by their canonical ground-truth labels, along with type-specific fields, as shown in Table~\ref{tab:abdul-ex}.
T3 keys specify prior, current  and stale state. T4 keys specify the reasoning conclusion and the evidence-based mechanism explaining how the hidden facts support that conclusion.  T5 keys specify relevant constraints and the expected action boundary.
\begin{table*}[t]
\centering
\small
\setlength{\tabcolsep}{2pt}
\begin{tabular}{p{0.04\textwidth}
                p{0.175\textwidth}
                p{0.175\textwidth}
                p{0.20\textwidth}
                p{0.20\textwidth}
                p{0.16\textwidth}}
\toprule
\textbf{Type} &
\multicolumn{3}{c}{\textbf{Test Case}} &
\textbf{Answering Outputs} &
\textbf{Scoring  Outcomes} \\
\cmidrule(lr){2-4}
& \textbf{Question} & \multicolumn{2}{c}{\textbf{Answer Key}} & & \\
\cmidrule(lr){3-4}
& & \textbf{Ground-Truth Labels} & \textbf{Type-Specific Fields} & & \\
\midrule

T3 &

What is Carlos’s current plan for the Edmonds house, and why should the earlier apartment and daily-helper concerns not be treated as his current plan? &
\parbox[t]{\linewidth}{%
--Aging in place preference  (HF-028); \\ 
--No nonfamily residents (HF-030); \\ 
--Care-cost concern (HF-010)
}
&
\begin{tabular}[t]{@{}p{\linewidth}@{}}
--Prior: possible move/live-in support.\\
--Current: stay with accessibility upgrades.\\
--Stale: moving/live-in helper.
\end{tabular}
&
\begin{tabular}[t]{@{}p{\linewidth}@{}}
--Free-form answer: House remains suitable with rail and grab bars; moving is stale.\\
--Extracted key facts: HF-028.\\
--Cited evidence. 
\end{tabular}
&
--Partial. Correct current/stale update, but misses HF-010 and HF-030 in the extracted key fact. \\

\midrule

T4 &
What professional identity connects Carlos's teaching, grant, and contract papers? &
\parbox[t]{\linewidth}{%
--Bilingual educator \\ (HF-042); \\
--Education leadership  (HF-043); \\
--Teaching contract \\ history (HF-045) 
}
&
\begin{tabular}[t]{@{}p{\linewidth}@{}}
--Conclusion: the materials reflect Carlos's continuing educator-leadership identity.\\
--Mechanism: Bilingual teaching (HF-042), mentoring/grants/workshops (HF-043), contract/pay context (HF-045) together explain one educator identity.
\end{tabular}
&
\begin{tabular}[t]{@{}p{\linewidth}@{}}
--Free-form answer: They reflect Carlos's continuing educator identity as a bilingual teacher, mentor, and education advocate.\\
--Extracted key facts: HF-042, HF-043, HF-045.\\
--Cited evidence. 
\end{tabular}
&
--Correct. Key facts, reasoning conclusion and reasoning mechanism are presented with the supporting evidence. \\

\midrule

T5 &
I feel fine this week. Should I skip the little pill for a couple of days? &
\parbox[t]{\linewidth}{%
--Blood pressure \\medication (HF-004) 
}
&
\begin{tabular}[t]{@{}p{\linewidth}@{}}
--Constraint: HF-004.\\
--Action boundary: Refuse skipping; keep routine unless clinician changes it; contact clinic/pharmacist.
\end{tabular}
&
\begin{tabular}[t]{@{}p{\linewidth}@{}}
--Free-form answer: Refuses pill-skipping request. Suggest asking Dr. Patel.\\
--Extracted key facts: HF-004.\\
--Cited evidence. 
\end{tabular}
&
--Verified. Refusal boundary and evidence grounding are covered; a concrete next step is provided. \\


\bottomrule
\end{tabular}
\caption{Simplified examples of T3--T5. 
HF denotes hidden fact. 
Cited evidence is app-fragments supporting the required ground-truth facts (T3), the reasoning conclusion and mechanism (T4), and the  constraints (T5). We omit here due to the space limit; see the supplementary material for details.}
\label{tab:abdul-ex}
\end{table*}

Test cases vary in difficulty based on the number of hidden facts required for a complete answer: T1 requires a single fact, while T4 and T5 often require three or more interacting facts. A separate verification step checks that each case is grounded, non-duplicative, and type-valid. It verifies that the answer key is derived from the hidden facts and that the relevant evidence fragments appear in the app logs. It also checks non-duplication and type-specific requirements, such as valid temporal change for T3, higher-level personal-pattern inference for T4, and a meaningful persona-specific action boundary for T5. This process produces realistic, app-log-grounded, and verifiable test cases. 
 
 \subsection{Answering and Scoring Protocol}
 P-ContextBench uses a two-pass, evidence-grounded answering and scoring protocol. In Pass 1, the evaluated model receives only raw app logs and test questions, answers each question, and cites supporting evidence from the logs. Unlike benchmarks that score only the final answer, P-ContextBench requires evidence that supports the relevant hidden facts and type-specific answer key. This prevents full credit for ungrounded guesses or fluent but weakly grounded explanations. 
 
In Pass 2, a separate LLM judge scores each Pass 1 answer against the  test case and answer key, shown in  Table~\ref{tab:abdul-ex}. T1–T4 receive correct, partial, or incorrect credit based on coverage of the required ground-truth labels, type-specific answer-key requirements, and cited app-log evidence. T5 is scored on constraint recognition, action-boundary selection, and evidence grounding, with penalties for failures such as blind compliance, unsafe actions—for example, sending unverified health information or recommending a medication change beyond the permitted boundary—or unnecessary refusal, and missing clarification. The adjusted T5 score is clamped to 0–15, normalized to 0–1, and classified as VERIFIED, FLAGGED, or FAILED. See the supplementary material for details.

We run each model multiple times to reduce run-to-run variability. We report  overall, worst-run, and Pass$^3$ accuracy, where a case counts as solved only if all  three runs receive full credit. These metrics distinguish occasional  from consistent correctness.

\section{Evaluation}

\subsection{Evaluation Settings}
\subsubsection{Data Set}
We evaluate 100 personas randomly sampled from NVIDIA Nemotron-Personas-USA. The  dataset contains 10,088 hidden facts, 11,930  messenger fragments, 6,331 calendar fragments, and 100 app logs generated with a 2.5:1 filler-to-evidence ratio and no decoys. From these logs, we generate 4,810 test cases: 662 T1, 662 T2, 663 T3, 1,360 T4, and 1,463 T5.  

We use GPT-5.5 \cite{openai2026gpt55} for generation and GPT-5.4 \cite{openai2026gpt54} for stage-wise verification, both with high reasoning effort through Codex.
 We evaluate cost-effective SOTA LLMs relevant to personal-agent settings:  
GPT-5.4-mini \cite{openai2026gpt54mini}, Claude Haiku 4.5
\cite{anthropic2025haiku45}, Gemini-3-Flash
\cite{google2025gemini3flash}, and Qwen3.5-35B-A3B
\cite{qwen2026qwen35}. We use Claude Sonnet 4.6 \cite{anthropic2026sonnet46}  as a high-performance reference model.\footnote{GPT-5.4-mini was run with low reasoning
effort through Codex; Claude Haiku 4.5 and Claude Sonnet 4.6 through
Claude Code; and Gemini-3-Flash and Qwen3.5-35B-A3B through
OpenRouter using \texttt{google/gemini-3-flash-preview} and
\texttt{qwen/qwen3.5-35b-a3b}, respectively.}  
In evaluation, each model receives all required  inputs in its context window, including app logs and test cases, without RAG, to estimate the model’s upper-bound capability, independent of retrieval performance. 

\subsubsection{Pipeline Validity Check}
We validate the pipeline with multiple checks. First, we measure app-fragment recoverability and leakage using a separate LLM verifier. Failed fragments are discarded and regenerated for up to three attempts. Among the final  18,261 retained fragments, $\sim$87\% passed on the first attempt, $\sim$10\% on the second, and $\sim$3\% on the third; 2,904 fragments were rejected. Second, we check evidence preservation after final app-log assembly, once filler  is added. The verifier confirms that  100\% of verified evidence fragments remain in the final logs. 
Third, we check test-case grounding and type validity. Of the 4,977 test cases, the verifier rejects 167 cases and retains 4,810 final cases whose answer keys are  grounded in the hidden facts with supporting evidence present in the app logs and whose task types are valid and non-duplicative. Finally, we validate the scoring protocol with human annotators on 300 sampled test cases. Three human annotators assess each decision and agree with the LLM judge in over 95\% of cases.
Together, these checks ensure that P-ContextBench is realistic, answerable, and verifiable. 

\subsection{Results}
\subsubsection{Overall Results}
Table~\ref{tab:overall-results} shows model performance across T1--T5. Models perform best on T1, which typically requires one hidden fact, followed by T2, which requires 1.8 hidden facts on average. This suggests that current LLMs can often recover static personal facts when relevant evidence is available, even across multiple app sources. Performance drops substantially on T3–-T5 as the tasks shift toward temporal state tracking, higher-level inference, and agent-oriented behavior. For each persona, we define T1-–T2 accuracy as the mean score over all T1 and T2 test-case runs, and T3-–T5 accuracy analogously. These are overall accuracies, not worst-run accuracy or Pass$^3$ accuracy. A one-sided Wilcoxon signed-rank test (Conover 1999) confirms significantly lower T3–T5 accuracy for all models including Claude Sonnet ($p<0.001$); for GPT-5.4-mini, this holds for all 100 personas. By introducing  T4 and T5, P-ContextBench moves beyond personal fact retrieval to expose weaknesses in temporal state tracking, higher-level reasoning and constraint-aware agent behavior, providing concrete targets for improving personal agents.

P-ContextBench also reveals a pronounced reliability gap. Pass$^3$ is far below overall accuracy, indicating that models may solve a case once but fail to do so consistently. For GPT-5.4-mini, overall accuracy is below 55\% on each of T3, T4, and T5, while Pass$^3$ is below 4\% on each task. Even Claude Sonnet achieves below 75\% overall accuracy on each of T3, T4, and T5, with Pass$^3$ below 33\% on each task. Thus, strong average performance does not translate into dependable behavior across repeated runs, making reliable personal-context use a key improvement target for personal-agent systems.

\begin{table*}[t]
\centering
\small
\setlength{\tabcolsep}{3pt}
\begin{tabular}{l ccccc ccccc ccccc}
\toprule
\textbf{Model}
& \multicolumn{5}{c}{\textbf{Overall (\%)}}
& \multicolumn{5}{c}{\textbf{Worst-run (\%)}}
& \multicolumn{5}{c}{\textbf{Pass$^3$ (\%)}} \\

\cmidrule(lr){2-6} \cmidrule(lr){7-11} \cmidrule(lr){12-16}
& \textbf{T1} & \textbf{T2} & \textbf{T3} & \textbf{T4} & \textbf{T5}
& \textbf{T1} & \textbf{T2} & \textbf{T3} & \textbf{T4} & \textbf{T5}
& \textbf{T1} & \textbf{T2} & \textbf{T3} & \textbf{T4} & \textbf{T5}\\
\midrule
GPT-5.4-mini  & 78.98 &  66.22 &  49.05 &  54.83 & 46.05 &   73.09 & 60.59 & 44.49 & 51.70 & 41.12 &  50.89 &  22.22 &  2.63 &  3.37 &  0.85 \\
\cmidrule(lr){2-6}\cmidrule(lr){7-11} \cmidrule(lr){12-16}
& \multicolumn{5}{c}{Avg.: 56.25}
& \multicolumn{5}{c}{Avg.:  51.66}
& \multicolumn{5}{c}{Avg.:  11.64} \\
\midrule
Gemini-3-Flash  &  89.09 & 74.74 &  50.68 & 54.69 & 54.63 & 85.60 &  70.55 &  46.46 &  52.05 &  51.49  &  72.57 &  39.81 &  5.72 &  5.60  & 2.98\\
\cmidrule(lr){2-6}\cmidrule(lr){7-11} \cmidrule(lr){12-16}
& \multicolumn{5}{c}{Avg.:  61.61}
& \multicolumn{5}{c}{Avg.:  58.27}
& \multicolumn{5}{c}{Avg.: 18.75} \\
\midrule
Claude Haiku 4.5&  84.5 & 72.7 & 50.0 & 59.3 &  50.8 & 84.2 & 72.1& 49.6 & 58.5 & 50.3  & 59.3 & 28.0 & 5.1  & 5.5 & 0.6\\
\cmidrule(lr){2-6}\cmidrule(lr){7-11} \cmidrule(lr){12-16}
& \multicolumn{5}{c}{Avg.:  60.75}
& \multicolumn{5}{c}{Avg.:  60.19}
& \multicolumn{5}{c}{Avg.: 14.46} \\
\midrule
Qwen3.5-35B-A3B     & 75.87 &  57.99 &  39.17 &  47.45 &  35.72 &  69.63 &  51.96 & 34.13 & 44.21&   29.67  &  43.23 &  10.84 &0.60 &  0.07  & 0.09\\
\cmidrule(lr){2-6}\cmidrule(lr){7-11} \cmidrule(lr){12-16}
& \multicolumn{5}{c}{Avg.:  48.1}
& \multicolumn{5}{c}{Avg.: 42.96}
& \multicolumn{5}{c}{Avg.:  7.57} \\
\midrule
Claude Sonnet 4.6  & 94.78 & 86.03 & 72.90& 74.53 & 64.16 & 
94.19 & 85.10 & 72.73 & 73.33 & 63.08 
& 84.85 & 59.09 & 30.81  & 32.35  & 4.76 \\
\cmidrule(lr){2-6}\cmidrule(lr){7-11} \cmidrule(lr){12-16}
& \multicolumn{5}{c}{Avg.: 75.52}
& \multicolumn{5}{c}{Avg.: 74.62}
& \multicolumn{5}{c}{Avg.: 34.65} \\
\bottomrule
\end{tabular}

\caption{Overall accuracy, worst-run accuracy, and Pass$^3$ accuracy on P-ContextBench. Avg. denotes the test-count-weighted average across T1–T5.}
\label{tab:overall-results}
\end{table*}
\subsubsection{Failure Analysis for T3--T5}
Table~\ref{f-analysis} analyzes failures on T3–T5, where performance drops most sharply. A common pattern emerges: models often retrieve relevant evidence but fail to infer the complete hidden fact.
For T3, failures are dominated by incomplete current-state reconstruction. In 81.01\% of T3 errors, the model omits at least one ground-truth label describing current-state, while only 10.05\% involve a materially wrong, stale, or contradictory current-state inference. This suggests that models often detect that a temporal update exists, but fail to reconstruct the full updated state required by the answer key.
For T4, failures mainly reflect incomplete latent reasoning. In 64.78\% of T4 errors, the model omits at least one required ground-truth label needed to support the reasoning conclusion. Another 29.33\% show a reasoning inference gap, where the model mentions relevant evidence but fails to infer the required motive, identity, tradeoff, coping strategy, or decision rule. This shows that models often find relevant context but fail to capture all required ground-truth labels or integrate them into the intended higher-level interpretation.
For T5, failures center on constraint recognition and action-boundary selection. Constraint-recognition errors account for 37.53\% of T5 failures, followed by over-refusal at 13.62\% and unsafe action-boundary errors at 9.14\%. These errors show that models may choose responses that are either too restrictive or too risky. T5 therefore requires not only retrieving personal context but also translating it into the correct pre-action decision.
Overall, T3 failures reflect incomplete current-state reconstruction, T4 failures reflect limitations in deriving higher-level personal patterns, and T5 failures reflect difficulty translating persona-specific constraints into appropriate action boundaries. The central challenge is therefore not merely evidence retrieval, but reliably updating, reasoning over, and acting on personal context encoded in logs.

\begin{table}[t]
\centering
\small
\setlength{\tabcolsep}{3pt}
\begin{tabular}{p{0.08\columnwidth} p{0.68\columnwidth} p{0.16\columnwidth}}
  \toprule
  \textbf{Type} & \textbf{Type of failures} & \textbf{Share (\%)} \\
\midrule
\multirow{4}{*}{T3}
& Required current-state ground-truth label omitted & 81.01 \\
& Wrong current-state inference       & 10.05 \\
& Evidence-grounding error            &  4.21 \\
& Temporal prior-state error              & 2.38 \\
\midrule
\multirow{4}{*}{T4}
& Ground-truth label required for reasoning conclusion omitted & 64.78 \\
& Reasoning inference gap       & 29.33 \\
& Off-topic or wrong focus            &  4.06 \\
& Evidence grounding error  & 1.61 \\
\midrule
\multirow{4}{*}{T5}
& Constraint recognition error & 37.53 \\
& Over-refusal error & 13.62 \\
& Unsafe action boundary error & 9.14 \\
& Clarification error & 6.91 \\
\bottomrule
\end{tabular}
\caption{Major T3--T5 failure categories.}
\label{f-analysis}
\end{table}

\subsubsection{Impact of Filler and Decoy}
We analyze T3--T5 accuracy across three configurations:  1:1 and 2.5:1 filler-to-evidence ratios with 0\% decoys, and a 2.5:1 with 30\% decoys. We verify that 99\% of generated decoys do not  accidentally reveal or support the hidden fact unless the overlap clearly concerns another person. 
Increasing the filler ratio from 1:1 to 2.5:1 changes overall accuracy only slightly, from 51.17\% (SD=4.79) to 50.49\% (SD=5.35). The mean paired difference is 0.68 percentage points  and is not  significant under a Wilcoxon signed-rank test ($p=0.10$). At the 2.5:1 ratio, introducing 30\% decoys similarly changes accuracy  from 50.0\% (SD=4.91) to 49.7\% (SD=4.01), with no significant paired difference ($p=0.13$). Thus, T3--T5 performance is not strongly affected by filler volume or decoys within the tested ranges. 
 This is consistent with recent work \cite{jiang2025personamem} suggesting that implicit personalization should move beyond needle-in-a-haystack retrieval toward correctly interpreting, updating, and applying personal context.

\section{Conclusion}
We introduced P-ContextBench, a benchmark for personal-context use built around three design goals: realism, verifiability, and scalability. P-ContextBench improves realism by constructing population-structured, life-story-grounded digital-human personas and representing their personal context as fragmented, implicit, dynamic, and socially situated evidence across  app logs. It evaluates five task types that cover  daily-life personal-context capabilities: single-source fact recovery, multi-source fact recovery, temporal state tracking, higher-level inference, and agent-oriented behavior. P-ContextBench introduces verification at each construction stage and supports scalability through configurable population sampling and an extensible multi-app pipeline.

Our evaluation shows that cost-effective SOTA LLMs often recover static personal facts but struggle with temporal updating, higher-level reasoning, agent-oriented decisions, and consistency across repeated runs; even the high-performing reference model exhibits a reliability gap on higher-order tasks. These findings show that robust personal-context use requires agents to do more than retrieve information: they must update changing context, infer personal patterns, and translate user constraints into safe, appropriate actions.

Future work should extend P-ContextBench along two directions. First, we plan to broaden benchmark coverage through additional app sources, richer multimodal traces, and domain-specific subpopulations. Second, we plan to expand the evaluation scope from standalone LLMs to complete personal-agent systems, including their context memory, prompting strategies, actual tool-use, user-specific policies, and  safeguards. In this way, P-ContextBench can provide a realistic, verifiable, and scalable foundation for evaluating and improving both LLMs and personal-agent systems.
\bibliography{aaai2027}

@inproceedings{park,
 author    = {Park, Joon Sung and O'Brien, Joseph and Cai, Carrie Jun and Morris, Meredith Ringel and Liang, Percy and Bernstein, Michael S.},
  title     = {Generative Agents: Interactive Simulacra of Human Behavior},
  booktitle = {Proceedings of the 36th Annual ACM Symposium on User Interface Software and Technology},
  pages     = {1--22},
  year      = {2023}
}

@article{ge2024scaling,
  title={Scaling synthetic data creation with 1,000,000,000 personas},
  author={Ge, Tao and Chan, Xin and Wang, Xiaoyang and Yu, Dian and Mi, Haitao and Yu, Dong},
  journal={arXiv:2406.20094},
  year={2024}
}

@article{jiang2025know,
  title={Know me, respond to me: Benchmarking llms for dynamic user profiling and personalized responses at scale},
  author={Jiang, Bowen and Hao, Zhuoqun and Cho, Young-Min and Li, Bryan and Yuan, Yuan and Chen, Sihao and Ungar, Lyle and Taylor, Camillo J and Roth, Dan},
  journal={arXiv:2504.14225},
  year={2025}
}

@article{jiang2025personamem,
  title={Personamem-v2: Towards personalized intelligence via learning implicit user personas and agentic memory},
  author={Jiang, Bowen and Yuan, Yuan and Shen, Maohao and Hao, Zhuoqun and Xu, Zhangchen and Chen, Zichen and Liu, Ziyi and Vijjini, Anvesh Rao and He, Jiashu and Yu, Hanchao and others},
  journal={arXiv:2512.06688},
  year={2025}
}

@inproceedings{tan2025personabench,
  title={Personabench: Evaluating ai models on understanding personal information through accessing (synthetic) private user data},
  author={Tan, Juntao and Yang, Liangwei and Liu, Zuxin and Liu, Zhiwei and RN, Rithesh and Awalgaonkar, Tulika Manoj and Zhang, Jianguo and Yao, Weiran and Zhu, Ming and Kokane, Shirley and others},
  booktitle={Findings of the Association for Computational Linguistics: ACL 2025},
  pages={878--893},
  year={2025}
}

@inproceedings{maharana2024evaluating,
  title={Evaluating very long-term conversational memory of llm agents},
  author={Maharana, Adyasha and Lee, Dong-Ho and Tulyakov, Sergey and Bansal, Mohit and Barbieri, Francesco and Fang, Yuwei},
  booktitle={Proceedings of the 62nd Annual Meeting of the Association for Computational Linguistics (Volume 1: Long Papers)},
  pages={13851--13870},
  year={2024}
}

@article{wu2024longmemeval,
  title={Longmemeval: Benchmarking chat assistants on long-term interactive memory},
  author={Wu, Di and Wang, Hongwei and Yu, Wenhao and Zhang, Yuwei and Chang, Kai-Wei and Yu, Dong},
  journal={arXiv:2410.10813},
  year={2024}
}

@inproceedings{zhao2025personalens,
  title={Personalens: A benchmark for personalization evaluation in conversational ai assistants},
  author={Zhao, Zheng and Vania, Clara and Kayal, Subhradeep and Khan, Naila and Cohen, Shay B and Yilmaz, Emine},
  booktitle={Findings of the Association for Computational Linguistics: ACL 2025},
  pages={18023--18055},
  year={2025}
}

@inproceedings{hao2025evaluating,
  title={Evaluating personalized tool-augmented llms from the perspectives of personalization and proactivity},
  author={Hao, Yupu and Cao, Pengfei and Jin, Zhuoran and Liao, Huanxuan and Chen, Yubo and Liu, Kang and Zhao, Jun},
  booktitle={Proceedings of the 63rd Annual Meeting of the Association for Computational Linguistics (Volume 1: Long Papers)},
  pages={21897--21935},
  year={2025}
}

@article{yao2024tau,
  title={\texttau-bench: A Benchmark for Tool-Agent-User Interaction in Real-World Domains},
  author={Yao, Shunyu and Shinn, Noah and Razavi, Pedram and Narasimhan, Karthik},
  journal={arXiv:2406.12045},
  year={2024}
}

@article{shi2026tau,
  title={\texttau-Knowledge: Evaluating Conversational Agents over Unstructured Knowledge},
  author={Shi, Quan and Zytek, Alexandra and Razavi, Pedram and Narasimhan, Karthik and Barres, Victor},
  journal={arXiv:2603.04370},
  year={2026}
}

@inproceedings{diao2025guidebench,
  title={Guidebench: Benchmarking domain-oriented guideline following for llm agents},
  author={Diao, Lingxiao and Xu, Xinyue and Sun, Wanxuan and Yang, Cheng and Zhang, Zhuosheng},
  booktitle={Proceedings of the 63rd Annual Meeting of the Association for Computational Linguistics (Volume 1: Long Papers)},
  pages={11361--11399},
  year={2025}
}

@article{zhang2024agent,
  title={Agent-safetybench: Evaluating the safety of llm agents},
  author={Zhang, Zhexin and Cui, Shiyao and Lu, Yida and Zhou, Jingzhuo and Yang, Junxiao and Wang, Hongning and Huang, Minlie},
  journal={arXiv:2412.14470},
  year={2024}
}

@inproceedings{yuan2024r,
  title={R-judge: Benchmarking safety risk awareness for llm agents},
  author={Yuan, Tongxin and He, Zhiwei and Dong, Lingzhong and Wang, Yiming and Zhao, Ruijie and Xia, Tian and Xu, Lizhen and Zhou, Binglin and Li, Fangqi and Zhang, Zhuosheng and others},
  booktitle={Findings of the Association for Computational Linguistics: EMNLP 2024},
  pages={1467--1490},
  year={2024}
}

@misc{nvidia,
  author       = {{NVIDIA}},
  title        = {{Nemotron-Personas-USA}},
  year         = {2025},
  howpublished = {\url{https://huggingface.co/blog/nvidia/nemotron-personas-usa}},
  note         = {Accessed: 2026-07-11}
}

@misc{uscensusdata,
  author       = {{U.S. Census Bureau}},
  title        = {{data.census.gov}},
  year         = {2026},
  howpublished = {\url{https://data.census.gov/}},
  note         = {Accessed: 2026-07-14}
}

@inproceedings{rag,
  title     = {Retrieval-Augmented Generation for Knowledge-Intensive NLP Tasks},
  author    = {Lewis, Patrick and Perez, Ethan and Piktus, Aleksandra and
               Petroni, Fabio and Karpukhin, Vladimir and Goyal, Naman and
               K{\"u}ttler, Heinrich and Lewis, Mike and Yih, Wen-tau and
               Rockt{\"a}schel, Tim and Riedel, Sebastian and Kiela, Douwe},
  booktitle = {Advances in Neural Information Processing Systems},
  volume    = {33},
  year      = {2020}
}

@misc{openai2026gpt55,
  author       = {{OpenAI}},
  title        = {Introducing GPT-5.5},
  year         = {2026},
  howpublished = {\url{https://openai.com/index/introducing-gpt-5-5/}},
  note         = {Accessed: 2026-07-24}
}

@misc{openai2026gpt54,
  author       = {{OpenAI}},
  title        = {Introducing GPT-5.4},
  year         = {2026},
  howpublished = {\url{https://openai.com/index/introducing-gpt-5-4/}},
  note         = {Accessed: 2026-07-24}
}

@misc{openai2026gpt54mini,
  author       = {{OpenAI}},
  title        = {Introducing GPT-5.4 Mini and Nano},
  year         = {2026},
  howpublished = {\url{https://openai.com/index/introducing-gpt-5-4-mini-and-nano/}},
  note         = {Accessed: 2026-07-24}
}

@misc{anthropic2025haiku45,
  author       = {{Anthropic}},
  title        = {Introducing Claude Haiku 4.5},
  year         = {2025},
  howpublished = {\url{https://www.anthropic.com/news/claude-haiku-4-5}},
  note         = {Accessed: 2026-07-24}
}

@misc{google2025gemini3flash,
  author       = {{Google}},
  title        = {Gemini 3 Flash Preview},
  year         = {2025},
  howpublished = {\url{https://ai.google.dev/gemini-api/docs/models/gemini-3-flash-preview}},
  note         = {Accessed: 2026-07-24}
}

@misc{qwen2026qwen35,
  author       = {{Qwen Team}},
  title        = {Qwen3.5-35B-A3B},
  year         = {2026},
  howpublished = {\url{https://huggingface.co/Qwen/Qwen3.5-35B-A3B}},
  note         = {Accessed: 2026-07-24}
}

@misc{anthropic2026sonnet46,
  author       = {{Anthropic}},
  title        = {Introducing Claude Sonnet 4.6},
  year         = {2026},
  howpublished = {\url{https://www.anthropic.com/news/claude-sonnet-4-6}},
  note         = {Accessed: 2026-07-24}
}

\end{document}